\documentclass[sn-mathphys]{sn-jnl}

\usepackage{setspace}

\pdfoutput=1

\jyear{2021}%

\theoremstyle{thmstyleone}%
\theoremstyle{thmstyletwo}%

\theoremstyle{thmstylethree}%

\begin{document}
	
	\title[ ]{Evaluating robustness of support vector machines with the Lagrangian dual approach}
	
	
	\author[1]{\fnm{Yuting} \sur{Liu}}
	
	\author[1]{\fnm{Hong} \sur{Gu}}

	\author*[1]{\fnm{Pan} \sur{Qin}}\email{qp112cn@dlut.edu.cn}
	
	\affil[1]{\orgdiv{Faculty of Electronic Information and Electrical Engineering}, \orgname{Dalian University of Technology}, \orgaddress{\street{2 Linggong Road}, \city{Dalian}, \postcode{116024}, \state{Liaoning}, \country{China}}}

	
	\abstract{Adversarial examples bring a considerable security threat to support vector machines (SVMs), especially those used in safety-critical applications. Thus, robustness verification is an essential issue for SVMs, which can provide provable robustness against various kinds of adversary attacks. The evaluation results obtained through the robustness verification can provide a safe guarantee for the use of SVMs. The existing verification method does not often perform well in verifying SVMs with nonlinear kernels. To this end, we propose a method to improve the verification performance for SVMs with nonlinear kernels. We first formalize the adversarial robustness evaluation of SVMs as an optimization problem. Then a lower bound of the original problem is obtained by solving the Lagrangian dual problem of the original problem. Finally, the adversarial robustness of SVMs is evaluated concerning the lower bound. We evaluate the adversarial robustness of SVMs with linear and nonlinear kernels on the MNIST and Fashion-MNIST datasets. The experimental results show that the percentage of provable robustness obtained by our method on the test set is better than that of the state-of-the-art.
	}

	\keywords{Support vector machines, Adversarial robustness, Robustness verification, Lagrangian Duality, Subgradient method}
	
	
	
	\maketitle
	\section{Introduction}\label{sec1}
	
	Machine learning has been widely used in safety-critical fields such as autonomous driving \cite{rais2022enhanced}, medical \cite{cui2022big}, and network security \cite{rajadurai2020stacked}. Adversarial examples, designed by adding subtle perturbations to the input to fool the model, bring a considerable security threat to well-trained models  \cite{szegedy2014intriguing}. Adversarial examples degenerate the model's prediction performance because the assumption that the training and test data are independently distributed is broken \cite{dvijotham2018dual}. In this case, the prediction performance criteria on test datasets, such as accuracy, precision, and F1-score, cannot prove robustness for models. Thus, it is necessary to develop methods to evaluate the robustness of models to subtle perturbations on the input.	
	
	So far, many studies have evaluated the adversarial robustness of machine learning models, most of which are designed for artificial neural networks (ANNs) \cite{dvijotham2018dual,xiao2021improving, kurakin2019adversarial, wang2021adversarial, gupta2022improved, wong2020learning,zhang2022black, chen2021camdar, wang2020spanning, andriushchenko2020square,kim2021robust, li2022decision, chen2020hopskipjumpattack, guo2020low,athalye2018obfuscated, uesato2018adversarial,zhu2021attack, liao2022robustness, xue2022rnn, tsay2021partition, tjeng2019evaluating,jia2020efficient, henzinger2021scalable, song2021qnnverifier, katz2019marabou, amir2021smt,wong2018provable,  raghunathan2018certified,gehr2018ai2, liu2020abstract, singh2019abstract, li2019analyzing, urban2020perfectly,ruan2018reachability, weng2018towards, latorre2020lipschitz}. The concept of adversarial examples was initially proposed by Szegedy et al. while exploring the interpretability of ANNs \cite{szegedy2014intriguing}. Szegedy et al. found that adding specific perturbed image samples could easily fool deep neural networks. This finding has spurred research into evaluating the adversarial robustness of ANNs. There are two approaches for adversarial robustness evaluation. One approach focuses on adversarial attacks, such as the gradient-based methods \cite{xiao2021improving, kurakin2019adversarial, wang2021adversarial, gupta2022improved, wong2020learning}, the score-based methods \cite{zhang2022black, chen2021camdar, wang2020spanning, andriushchenko2020square}, and the decision-based methods \cite{kim2021robust, li2022decision, chen2020hopskipjumpattack, guo2020low}. Those studies use adversarial attacks to obtain minimal perturbation added to sample features to evaluate the robustness of ANNs. However, the research in \cite{athalye2018obfuscated, uesato2018adversarial} has shown that the robustness evaluation results obtained in a specific adversarial attack cannot usually provide robustness guarantees for other attacks. The other approach is robustness verification methods, which are developed by the mixed integer linear programming \cite{zhu2021attack, liao2022robustness, xue2022rnn, tsay2021partition, tjeng2019evaluating}, the satisfiability modulo theory \cite{jia2020efficient, henzinger2021scalable, song2021qnnverifier, katz2019marabou, amir2021smt},
	the duality in optimization theory \cite{wong2018provable, dvijotham2018dual, raghunathan2018certified}, the abstract interpretation \cite{gehr2018ai2, liu2020abstract, singh2019abstract, li2019analyzing, urban2020perfectly}, and bounding the local Lipschitz constant \cite{ruan2018reachability, weng2018towards, latorre2020lipschitz}. The robustness verification method investigates how the outputs of models change under a given range of perturbations. For this reason, the robustness verification method can provide provable robustness for ANNs against various kinds of adversary attacks. 
	
	Note that the problem of lack of adversarial robustness is not unique to ANNs. Compared with the limitation of ANNs \cite{cervantes2020comprehensive}, support vector machines (SVMs) have a more simple structure and solid mathematical theoretical foundation and are widely used in safety-critical fields \cite{biggio2014security}.	At present, there are relatively few studies on evaluating the adversarial robustness of SVMs. Meanwhile, most of these studies focus on adversarial attacks, such as \cite{biggio2013evasion, zhang2015adversarial,weerasinghe2021defending}. Thus, it is necessary to develop robustness verification methods to evaluate the adversarial robustness of SVMs. Note that a robustness verification method, named SAVer, is proposed by Ranzato et al. based on the abstract interpretation \cite{ranzato2019robustness}, which can provide provable robustness for SVMs against various kinds of adversary attacks. Specifically, SAVer uses an abstraction that combines interval domains \cite{cousot1977abstract} and reduced affine form domains \cite{ranzato2019robustness} to provide provable robustness for SVMs. However, for evaluating the robustness of SVMs with nonlinear kernels, the abstract nonlinear operations applied by SAVer on the interval domain and the reduced affine form domain lead to a loss of computational accuracy. The experimental results in \cite{ranzato2019SAVer} show that SAVer does not often perform well in verifying SVMs with nonlinear kernels.
	
	To this end, we propose a robustness verification method called SVM dual verifier (SDVer), based on the Lagrangian duality. We first formulate the optimization problem of various kernels in SVMs into feedforward neural network representations. Then, considering that the solution to the original problem is generally NP-hard, we transform the object to be solved from the original problem into the Lagrangian dual problem of the original problem. By using the subgradient method to solve the Lagrangian dual problem, a lower bound of the original optimization problem is obtained. The adversarial robustness of SVMs is evaluated according to the lower bound. In this way, we summarize the adversarial robustness evaluation of SVMs as an optimization problem. Finally, we compare our method with the state-of-the-art SVMs robustness verifier SAVer on the MNIST \cite{lecun1998gradient} and Fashion-MNIST \cite{xiao2017fashion} datasets. The results indicate that when the kernel function is linear, the percentage of provable robustness obtained by our method is consistent with that obtained by SAVer. When the kernel function is nonlinear, the percentage of provable robustness obtained by our method is better than that of SAVer.
	
	Our main contributions are as follows:
	
	We proposed SDVer to evaluate the adversarial robustness of SVMs. The method is based on the Lagrangian duality considering the kernels used in SVMs. The method is proposed based on the idea of robustness verification, which can provide provable robustness for SVMs against various kinds of adversarial attacks. With the proposed SDVer, we significantly improve the verification performance for SVMs with nonlinear kernels in experiments.
	
	The rest of the paper is organized as follows. Section 2 introduces some preliminaries. Section 3 is a detailed introduction to our proposed method for the adversarial robustness evaluation of SVMs. In Section 4, we conduct experimental evaluation and analysis. Finally, Section 5 is the conclusion of the paper.
	
	\section{Preliminaries}\label{sec2}
	\subsection{Support vector machines}\label{subsec2.1}
	To elaborate on the robustness verification method of SVMs, we first give a brief introduction to SVMs. Given a training dataset $D=\left\{\left(\boldsymbol{x}_{i}, y_{i}\right)\mid i=1,\ldots,N \right\} \subseteq \mathbb{R}^{n} \times \{-1,+1\}$,
	training SVM is to find a hyperplane equation $f(\boldsymbol{x})$ that divides training samples into different classes in the sample space. We denote the trained SVM binary classifier as $\hat{y} = C_{\{+1,-1\}}(\boldsymbol{x})$, with $\hat{y}$ being the prediction of $y$. 
	\begin{align}
		&\hat{y} = C_{\{+1,-1\}}(\boldsymbol{x}) = \operatorname{sign}\left(f(\boldsymbol{x})\right), \\
		&f(\boldsymbol{x}) =\sum_{i=1}^{m} \alpha_{i} y_{i} \kappa\left(\boldsymbol{x}, \boldsymbol{x}_{i}\right)+b,
	\end{align}
	where $\alpha_{i}\in \mathbb{R}$ and $b\in \mathbb{R}$ are the parameters obtained after training, 
	$\left\{\left(\boldsymbol{x}_{i}, y_{i}\right) \mid i=1,2,\ldots,m \right\} \subseteq \mathbb{R}^{n} \times \{-1,+1\}$ represent support vectors,
	and $\kappa\left(\boldsymbol{x}_{i}, \boldsymbol{x}_{j}\right):\mathbb{R}^{n} \times \mathbb{R}^{n} \to \mathbb{R} $ is a kernel function. This paper mainly analyzes the adversarial robustness of SVMs for the most commonly used kernel functions listed in Table~\ref{tab1}. 
	\begin{table}[h]
		\begin{center}
			\begin{minipage}{260pt}
				\caption{Most commonly used kernel functions}
				\renewcommand\arraystretch{1.25}
				\begin{tabular}{@{}ll@{}}
					\toprule
					Name & Expression\\
					\midrule
					linear kernel & $\kappa\left(\boldsymbol{x}_{i}, \boldsymbol{x}_{j}\right)=\boldsymbol{x}_{i}^{\top} \boldsymbol{x}_{j}$ \\
					
					polynomial kernel & $\kappa\left(\boldsymbol{x}_{i}, \boldsymbol{x}_{j}\right)=\left(\boldsymbol{x}_{i}^{\top} \boldsymbol{x}_{j} + c\right)^{d}, \quad d \geqslant 1$\\
					
					sigmoid kernel & $\kappa\left(\boldsymbol{x}_{i}, \boldsymbol{x}_{j}\right)=\tanh \left(\beta \boldsymbol{x}_{i}^{\top} \boldsymbol{x}_{j}+\theta\right), \quad \beta>0, \theta<0$\\
					Gaussian kernel (RBF) & $\kappa\left(\boldsymbol{x}_{i}, \boldsymbol{x}_{j}\right)=\exp\left( \mid\mid\boldsymbol{x}_{i}-\boldsymbol{x}_{j}\mid\mid^{2}/2 \sigma^{2}\right), \quad \sigma>0$ \\
					
					\botrule
				\end{tabular} \label{tab1}
			\end{minipage}
		\end{center}
	\end{table}

	\subsection{Robustness verification goal of SVMs}\label{subsec2.2}
	We first define the adversarial region to describe the robustness verification goal of SVMs. The adversarial region is often defined by an $l_{\infty}$-norm ball. Formally, taking the input $\boldsymbol{x}$ as the center of the ball and the perturbation $\delta>0$ as the radius, the adversarial region denotes by $ P_{\delta}^{\infty}(\boldsymbol{x}) =\left\{\boldsymbol{x}^{\prime} \in \mathbb{R}^{n} \mid\left\|\boldsymbol{x}^{\prime}-\boldsymbol{x}\right\|_{\infty} \leq \delta\right\}=\left\{\boldsymbol{x}^{\prime} \in \mathbb{R}^{n} \mid \forall i , {x}_{i}^{\prime} \in\left[{x}_{i}-\delta, {x}_{i}+\delta\right] \right\}$,
	where ${x}_{i} \text{ is the $i$-th element of } \boldsymbol{x}$.
	Then we define the robustness verification goal of SVMs in the adversarial region. Given a trained SVM model $C_{\{+1,-1\}}(\boldsymbol{\cdot})$,  a test example $(\boldsymbol{x}, y)$, and a perturbation $\delta$, if we can verify that $C_{\{+1,-1\}}(\boldsymbol{x}^{\prime})= C_{\{+1,-1\}}(\boldsymbol{x})$ holds for $\boldsymbol{x}^{\prime} \in  P_{\delta}^{\infty}(\boldsymbol{x})$, then sample $\boldsymbol{x}$ is considered robust in its adversarial region.	
	We formulate the robustness evaluation of SVMs as an optimization problem (\ref{RVO}).
	If the optimal value of (\ref{RVO}) is larger than 0, then the sample is considered robust in its adversarial region. Otherwise, it is not robust in its adversarial region.
	\begin{subequations}\label{RVO}
		\begin{alignat}{2}
			&\min_{\boldsymbol{x}^{\prime}} &\quad & \hat{y} \cdot f(\boldsymbol{x}^{\prime}), \label{RVOa}\\
			&s.t. & & f(\boldsymbol{x}^{\prime}) = \sum\limits_{i=1}^{m} \alpha_{i} y_{i} k\left(\boldsymbol{x}^{\prime},\boldsymbol{x}_{i}\right) +b, \label{RVOb}\\
			&&&\boldsymbol{x}^{\prime} \in P_{\delta}^{\infty}(\boldsymbol{x}). \label{RVOc}
		\end{alignat}
	\end{subequations}

	\section{Methodology}\label{sec3}
	\subsection{Feedforward neural network representation of SVMs}\label{subsec3.1}
	Our method is developed under the framework of the verification method proposed in \cite{dvijotham2018dual}. Thus, we formulate the optimization problem (\ref{RVO}) as a feedforward neural network representation. Let $\boldsymbol{h}^{l}\left(\boldsymbol{x}\right)$ denotes the vector activation function for layer $l$, and  ${h}_{k}^{l}(x)$ denotes the $k$-th element of $\boldsymbol{h}^l\left(\boldsymbol{x}\right)$. When the kernel function is linear, (\ref{RVO}) can be expressed as follows:
	\begin{subequations}\label{LIN}
		\begin{alignat}{2}
			&\min_{\boldsymbol{x}^{\prime}} &\quad & \hat{y} \cdot f(\boldsymbol{x}^{\prime}), \label{LINa}\\
			&s.t. & & \boldsymbol{z}^{0} ={W^{0}}^{\top} \boldsymbol{x}^{\prime}, \label{LINb}\\
			&&& \boldsymbol{x}^{1} = \boldsymbol{h}^{0}\left(\boldsymbol{z}^{0}\right),	  \label{LINc}\\
			&&& f(\boldsymbol{x}^{\prime}) = {\boldsymbol{w}^{1}}^{\top} \boldsymbol{x}^{1}+ {b}^{1}, \label{LINd}\\
			&&& \boldsymbol{x}^{\prime} \in P_{\delta}^{\infty}(\boldsymbol{x}),	\label{LINe}\
		\end{alignat}
	\end{subequations}
	where $W^{0}=\left(\boldsymbol{x}_{1}, \ldots, \boldsymbol{x}_{m}\right) \in\mathbb{R}^{n\times m} $, $\boldsymbol{h}^{0}\left(\boldsymbol{x}\right)=\boldsymbol{x}\in\mathbb{R}^{m}$ is a vector identity function, $\boldsymbol{w}^{1}=\left(\alpha_1 y_1, \ldots, \alpha_m y_m \right)^{\top} \in\mathbb{R}^{m}$, and ${b}^1=b\in\mathbb{R}$.  
	According to the classical formula of ANN \cite{goodfellow2016deep}, (\ref{LINb})-(\ref{LINd}) can be considered as a representation of a single hidden layer neural network, with $\boldsymbol{h}^{0}\left(\boldsymbol{x}\right) = \boldsymbol{x}$ as its activation function. When the kernel function is polynomial, (\ref{RVO}) can be expressed by (\ref{poly}).	
	\begin{subequations}\label{poly}
		\begin{alignat}{2}
			&\min_{\boldsymbol{x}^{\prime}} &\quad & \hat{y} \cdot f(\boldsymbol{x}^{\prime}),	\label{POLYa}\\
			&s.t. & & \boldsymbol{z}^{0} = {W^{0}}^{\top} \boldsymbol{x}^{\prime}+\boldsymbol{b}^{0}, \label{POLYb}\\
			& & &\boldsymbol{x}^{1} =\boldsymbol{h}^{0}\left(\boldsymbol{z}^{0}\right), \label{POLYc}\\
			& & &f(\boldsymbol{x}^{\prime}) = {\boldsymbol{w}^{1}}^{\top} \boldsymbol{x}^{1}+{b}^{1},	\label{POLYd}\\
			& & &\boldsymbol{x}^{\prime} \in P_{\delta}^{\infty}(\boldsymbol{x}),	\label{POLYe}
		\end{alignat}
	\end{subequations}		
	where $W^{0}=\left(\boldsymbol{x}_{1}, \ldots, \boldsymbol{x}_{m}\right) \in\mathbb{R}^{n\times m}$, $\boldsymbol{b}^0=\left(c, \ldots, c\right)^{\top}\in\mathbb{R}^{m}$, $\boldsymbol{h}^{0}\left(\boldsymbol{x}\right)=\boldsymbol{x}^d \in\mathbb{R}^{m}$, $\boldsymbol{w}^{1}=\left(\alpha_1 y_1, \ldots, \alpha_m y_m\right)^{\top}\in\mathbb{R}^{m}$, and ${b}^1=b\in\mathbb{R}$. Similar to the linear kernel, (\ref{POLYb})-(\ref{POLYd}) can be considered as a representation of a single hidden layer neural network, with $\boldsymbol{h}^{0}\left(\boldsymbol{x}\right) = \boldsymbol{x}^d$ as its activation function. When the kernel function is sigmoid, (\ref{RVO}) can be expressed by (\ref{sig}).	
	\begin{subequations}\label{sig}
		\begin{alignat}{2}
			&\min_{\boldsymbol{x}^{\prime}} &\quad& \hat{y} \cdot  f(\boldsymbol{x}^{\prime}), \label{SIGa}\\
			&s.t. & & \boldsymbol{z}^{0} = {W^{0}}^{\top} \boldsymbol{x}^{\prime}+\boldsymbol{b}^{0},	\label{SIGb}\\
			&&&\boldsymbol{x}^{1} = \boldsymbol{h}^{0}\left(\boldsymbol{z}^{0}\right), \label{SIGc}\\
			&&&f(\boldsymbol{x}^{\prime}) = {\boldsymbol{w}^{1}}^{\top} \boldsymbol{x}^{1}+{b}^{1},	\label{SIGd}\\
			&&&\boldsymbol{x}^{\prime} \in P_{\delta}^{\infty}(\boldsymbol{x}),	\label{SIGe}
		\end{alignat}
	\end{subequations}
	where $W^{0}=\left(\beta\boldsymbol{x}_{1}, \ldots, \beta\boldsymbol{x}_{m}\right)\in\mathbb{R}^{n\times m}$, $\boldsymbol{b}^0=\left(\theta, \ldots, \theta\right)^{\top}\in\mathbb{R}^{m}$, $\boldsymbol{h}^{0}\left(\boldsymbol{x}\right)=\text{tanh}(\boldsymbol{x})\in\mathbb{R}^{m}$, $\boldsymbol{w}^{1}=\left(\alpha_1 y_1, \ldots, \alpha_m y_m\right)^{\top}\in\mathbb{R}^{m}$, and ${b}^1=b\in\mathbb{R}$.	Similar to the linear and polynomial kernel, (\ref{SIGb})-(\ref{SIGd}) can be viewed as a representation of a single hidden layer neural network, with $\boldsymbol{h}^{0}\left(\boldsymbol{x}\right)=\text{tanh}(\boldsymbol{x})$ as its activation function. When the kernel function is RBF, (\ref{RVO}) can be represented as (\ref{gauss}).
	\begin{subequations}\label{gauss}
		\begin{alignat}{2}
			&\min_{\boldsymbol{x}^{\prime}} &\quad & \hat{y} \cdot f(\boldsymbol{x}^{\prime}), \label{GAUSSa}\\
			&s.t. &&\boldsymbol{z}^{0} = {W^{0}}^{\top} \boldsymbol{x}^{\prime}+\boldsymbol{b}^{0}, \label{GAUSSb}\\
			&&&\boldsymbol{x}^{1} =\boldsymbol{h}^{0}\left(\boldsymbol{z}^{0}\right),	\label{GAUSSc}\\
			&&&\boldsymbol{z}^{1} ={ W^{1}}^{\top} \boldsymbol{x}^{1}, \label{GAUSSd}\\
			&&&\boldsymbol{x}^{2} =\boldsymbol{h}^{1}\left(\boldsymbol{z}^{1}\right),	\label{GAUSSe}\\
			&&& f(\boldsymbol{x}^{\prime}) = {\boldsymbol{w}^{2}}^{\top} \boldsymbol{x}^{2}+{b}^{2},	\label{GAUSSf}\\
			&&&\boldsymbol{x}^{\prime}  \in P_{\delta}^{\infty}(\boldsymbol{x}),	\label{GAUSSg}
		\end{alignat}
	\end{subequations}
	where $W^{0}=\left({I}_{n}, \ldots, {I}_{n}\right)\in\mathbb{R}^{n\times mn}$ with ${I}_{n}\in \mathbb{R}^{n\times n}$ being an identity matrix, $\boldsymbol{b}^{0}=-\left(\boldsymbol{x}_{1} ,\ldots, \boldsymbol{x}_{m}\right)^{\top}\in\mathbb{R}^{mn}$, $\boldsymbol{h}^{0}\left(\boldsymbol{x}\right)=\boldsymbol{x}^2\in\mathbb{R}^{mn}$, $\boldsymbol{h}^{1}\left(\boldsymbol{x}\right)=e^{-\gamma \boldsymbol{x}}\in\mathbb{R}^{m}$, $\boldsymbol{w}^{2}=\left(\alpha_1 y_1, \ldots, \alpha_m y_m\right)^{\top}\in\mathbb{R}^{m}$, ${b}^{2}=b\in\mathbb{R}$, and  $W^{1}\in\mathbb{R}^{mn\times m}$ is as the following:
	$$
	W^{1}=\left(\begin{array}{cccc}
	\boldsymbol{1}_{n}& \boldsymbol{0}_{n} & \cdots & \boldsymbol{0}_{n} \\
	\boldsymbol{0}_{n} & \boldsymbol{1}_{n}&\cdots & \boldsymbol{0}_{n} \\
	\vdots  & \vdots &\vdots &\vdots  \\
	\boldsymbol{0}_{n} & \cdots&\boldsymbol{0}_{n} & \boldsymbol{1}_{n}
	\end{array}\right),
	$$
	with $\boldsymbol{1}_{n}\in\mathbb{R}^{n}$ being an $n$-dimensional vector with all one elements and $\boldsymbol{0}_{n}\in\mathbb{R}^{n}$ being an $n$-dimensional vector with all zero elements. (\ref{GAUSSb})-(\ref{GAUSSf}) can be considered as a representation of two hidden layer neural network.
	$\boldsymbol{h}^{0}\left(\boldsymbol{x}\right)=\boldsymbol{x}^2 \text{ and } \boldsymbol{h}^{1}\left(\boldsymbol{x}\right)=e^{-\gamma \boldsymbol{x}}$ are the activation functions of the first and second layer networks, respectively.
	
	To obtain a more general feedforward neural network representation of (\ref{RVO}), we set $\boldsymbol{z}^{L} = f(\boldsymbol{x}^{\prime})$, $\boldsymbol{x}^{0} = \boldsymbol{x}^{\prime}$, $\boldsymbol{\hat{y}} = \hat{y}$ and denote the optimal value of the optimization problem by $p^{*}$ to obtain the following expression.
	\begin{subequations}\label{NNR}
		\begin{alignat}{2}
			&p^{*} = &\quad&\min_{ \boldsymbol{x}^{0}} \boldsymbol{\hat{y}}^{\top} \boldsymbol{z}^{L}, \label{NNRa}\\
			&s.t. && \boldsymbol{x}^{l+1}=\boldsymbol{h}^{l}\left(\boldsymbol{z}^{l}\right),\quad l=0,1, \ldots, L-1,  \label{NNRb}\\
			&&& \boldsymbol{z}^{l}={W^{l}}^{\top} \boldsymbol{x}^{l}+\boldsymbol{b}^{l}, \quad l=0,1, \ldots, L, \label{NNRc}\\
			&&&  \boldsymbol{x}^{0}\in P_{\delta}^{\infty}(\boldsymbol{x}). \label{NNRd}
		\end{alignat}
	\end{subequations}

	\subsection{Robustness verification based on the Lagrangian duality for SVMs}\label{subsec3.2}
	A general feedforward neural network representation of (\ref{RVO}) can be obtained from \ref{subsec3.1}, as shown in (\ref{NNR}). When the kernel function is sigmoid, the corresponding optimization problem is a type of sigmoid programming problem. The work \cite{udell2013maximizing} has proved that the solution to the sigmoid programming problem is NP-hard. Motivated by \cite{dvijotham2018dual}, we develop a robustness verification method based on the Lagrangian duality to solve (\ref{NNR}) as follows:
	
	The Lagrangian multipliers $\boldsymbol{\mu}^{l} \text{ and } \boldsymbol{\lambda}^{l}$ are introduced to relax the equations (\ref{NNRb}) and (\ref{NNRc}), then the Lagrangian dual problem of the original problem is obtained.
	\begin{align}\label{OP}
		\begin{split}
			&L^{*} = \max_{\boldsymbol{\mu},\boldsymbol{\lambda}}\; L(\boldsymbol{\mu},\boldsymbol{\lambda}), \\
			&L(\boldsymbol{\mu},\boldsymbol{\lambda}) = \min_{\boldsymbol{z}^{l},\boldsymbol{x}^{l}}\; \boldsymbol{\hat{y}}^{\top}\left({W^{L}}^{\top} \boldsymbol{x}^{L}+\boldsymbol{b}^{L}\right)
			+\sum_{l=0}^{L-1}\left(\boldsymbol{\mu}^{l}\right)^{\top}\left(\boldsymbol{z}^{l}- {W^{l}}^{\top} \boldsymbol{x}^{l}-\boldsymbol{b}^{l}\right) \\ &\qquad\qquad+\sum_{l=0}^{L-1}\left(\boldsymbol{\lambda}^{l}\right)^{\top}\left(\boldsymbol{x}^{l+1}-\boldsymbol{h}^{l}\left(\boldsymbol{z}^{l}\right)\right),\\
			&s.t.\quad \underline{\boldsymbol{z}}^{l} \leq \boldsymbol{z}^{l} \leq \overline{\boldsymbol{z}}^{l},\quad l=0, 1, \ldots, L-1, \\
			&\quad\quad\;\underline{\boldsymbol{x}}^{l}  \leq \boldsymbol{x}^{l} \leq \overline{\boldsymbol{x}}^{l},\quad l=0, 1,  \ldots, L, \\
			&\quad\quad\;\boldsymbol{x}^{0}\in P_{\delta}^{\infty}(\boldsymbol{x}),
		\end{split}
	\end{align}
	where $\underline{\boldsymbol{z}}^{l} \text{ and } \overline{\boldsymbol{z}}^{l}$ are lower and upper bounds of $\boldsymbol{z}^{l}$; $\underline{\boldsymbol{x}}^{l} \text{ and } \overline{\boldsymbol{x}}^{l}$ are lower and upper bounds of $\boldsymbol{x}^{l}$. The values of $\underline{\boldsymbol{z}}^{l}$ and $\overline{\boldsymbol{z}}^{l}$ can be calculated by (\ref{ZLa}) and (\ref{ZUb}), respectively.
	\begin{subequations}
		\begin{align}
			&\underline{\boldsymbol{z}}^{l}=\left[W^{l}\right]_{+} \underline{\boldsymbol{x}}^{l}+\left[W^{l}\right]_{-} \overline{\boldsymbol{x}}^{l}+\boldsymbol{b}^{l}, \label{ZLa}\\
			&\overline{\boldsymbol{z}}^{l}=\left[W^{l}\right]_{+} \overline{\boldsymbol{x}}^{l}+\left[W^{l}\right]_{-} \underline{\boldsymbol{x}}^{l}+\boldsymbol{b}^{l}, \label{ZUb}
		\end{align}	
	\end{subequations}
	where $\left[W^{l}\right]_{+}= \max\left(W^{l},0\right)$ and  $\left[W^{l}\right]_{-}= \min\left(W^{l},0\right)$.
	The values of $\underline{\boldsymbol{x}}^{l}$ and $\boldsymbol{\overline{x}}^{l}$ are calculated according to the specific expression of $\boldsymbol{h}$. The expression of $\boldsymbol{h}$ is related to the choice of kernels in Table~\ref{tab1}. When $\boldsymbol{h}$ is a monotonically increasing function, the upper and lower bounds of each dimension of $\boldsymbol{x}$ are computed as shown in (\ref{IXLa}) and (\ref{IXUb}).
	\begin{subequations}
		\begin{align}
			&\underline{x}_{k}^{l+1}={h}_{k}^{l}\left(\underline{z}_{k}^{l}\right),  \label{IXLa}\\
			&\overline{x}_{k}^{l+1}={h}_{k}^{l}\left(\overline{z}_{k}^{l}\right). \label{IXUb}
		\end{align}	
	\end{subequations}
	When $\boldsymbol{h}$ is a monotonically decreasing function, the upper and lower bounds of each dimension of $\boldsymbol{x}$ are calculated as shown in  (\ref{DXLa}) and (\ref{DXUb}).
	\begin{subequations}
		\begin{align}
			&\underline{x}_{k}^{l+1}={h}_{k}^{l}\left(\overline{z}_{k}^{l}\right),\label{DXLa}\\
			&\overline{x}_{k}^{l+1}= {h}_{k}^{l}\left(\underline{z}_{k}^{l}\right). \label{DXUb}
		\end{align}	
	\end{subequations}
	When $\boldsymbol{h}$ is a nonmonotonic function, the upper and lower bounds of each dimension of $\boldsymbol{x}$ are computed by (\ref{NIDXLa}) and (\ref{NIDXUb}).
	\begin{subequations}
		\begin{align}
			&\underline{x}_{k}^{l+1}=\left\{
			\begin{array}{ll}
				0,  & \underline{z}_{k}^{l} \le 0 \le \overline{z}_{k}^{l}, \\
				\min\left\{ {h}_{k}^{l}\left(\overline{z}_{k}^{l}\right),{h}_{k}^{l}\left(\underline{z}_{k}^{l}\right)\right\}, & \text{otherwise},
			\end{array}\right. \label{NIDXLa}\\
			&\overline{x}_{k}^{l+1}= \max\left\{ {h}_{k}^{l}\left(\overline{z}_{k}^{l}\right),{h}_{k}^{l}\left(\underline{z}_{k}^{l}\right)\right\}.	\label{NIDXUb}
		\end{align}	
	\end{subequations}
	According to \cite{ahuja1988network}, the following inequality holds.
	\begin{align}\label{NF}
		L(\boldsymbol{\mu},\boldsymbol{\lambda}) \le L^{*} \le p^* \le \boldsymbol{\hat{y}}^{\top} \boldsymbol{z}^{L}.
	\end{align}	
	Under the definition of robustness verification goal in \ref{subsec2.2},  if $p^{*}$ is larger than 0, SVM is considered robust in the adversarial region $P_{\delta}^{\infty}(\boldsymbol{x})$.
	If there exist $\boldsymbol{\mu}$ and $\boldsymbol{\lambda}$ for $L(\boldsymbol{\mu},\boldsymbol{\lambda}) > 0$, $p^{*}$ must be positive concerning (\ref{NF}).
	
	Then, under the condition of fixed $\boldsymbol{\mu}$ and $\boldsymbol{\lambda}$, $L(\boldsymbol{\mu},\boldsymbol{\lambda})$ can be decomposed into the following three optimization problems:
	\begin{subequations}\label{xz}
		\begin{align}
			&f_{l}\left( \boldsymbol{\mu}^{l}, \boldsymbol{\lambda}^{l-1}\right)= 
			\min_{\boldsymbol{x}^{l} \in\left[\underline{\boldsymbol{x}}^{l}, \overline{\boldsymbol{x}}^{l}\right]}\left(\boldsymbol{\lambda}^{l-1}-W^{l} \boldsymbol{\mu}^{l}\right)^{\top} \boldsymbol{x}^{l}-\left(\boldsymbol{b}^{l}\right)^{\top}\boldsymbol{\mu}^{l},
			\quad l=1,\cdots,L, \label{dtoa}\\
			&\tilde{f}_{l}\left( \boldsymbol{\mu}^{l}, \boldsymbol{\lambda}^{l}\right)=\min _{\boldsymbol{z}^{l} \in\left[\underline{\boldsymbol{z}}^{l}, \overline{\boldsymbol{z}}^{l}\right]} \left({\boldsymbol{\mu}^{l}}\right)^{\top} \boldsymbol{z}^{l}-\left(\boldsymbol{\lambda}^{l}\right)^{\top} \boldsymbol{h}^{l}\left(\boldsymbol{z}^{l}\right),
			\quad l=0,\cdots,L-1,  \label{dtob}\\
			&f_{0}\left(\boldsymbol{\mu}^{0}\right)=\min_{\boldsymbol{x}^{0}\in P_{\delta}^{\infty}(\boldsymbol{x})}\left(-W^{0} \boldsymbol{\mu}^{0}\right)^{\top} \boldsymbol{x}^{0}-\left(\boldsymbol{b}^{0}\right)^{\top} \boldsymbol{\mu}^{0} , \label{dtoc}
		\end{align}
	\end{subequations}
	where (\ref{dtoa}) can be solved by (\ref{f1}). (\ref{dtob})  is essentially a one-dimensional optimization problem, as shown in (\ref{f2}). The optimal solution can be easily obtained according to the specific form of $h$. (\ref{dtoc}) is solved in the same way as (\ref{f1}).
	\begin{subequations}
		\begin{align}
			&f_{l}\left(\boldsymbol{\mu}^{l}, \boldsymbol{\lambda}^{l-1}\right)= 
			{\left[\boldsymbol{\lambda}^{l-1}-W^{l} \boldsymbol{\mu}^{l}\right]_{+}^{\top} \overline{\boldsymbol{x}}^{l}} 
			+\left[\boldsymbol{\lambda}^{l-1}- W^{l} \boldsymbol{\mu}^{l}\right]_{-}^{\top} \underline{\boldsymbol{x}}^{l}-\left(\boldsymbol{b}^{l}\right)^{\top} \boldsymbol{\mu}^{l}, \label{f1}\\
			&\tilde{f}_{l, k}\left(\mu_{k}^{l}, \lambda_{k}^{l}\right)=\min _{z_{k}^{l} \in\left[\underline{z} _{k}^{l}, \overline{z}_{k}^{l}\right]} \mu_{k}^{l} z_{k}^{l}-\lambda_{k}^{l} h_{k}^{l}\left(z_{k}^{l}\right). \label{f2}
		\end{align}
	\end{subequations}
	
	After solving (\ref{dtoa}), (\ref{dtob}) and (\ref{dtoc}) , using the subgradient method \cite{boyd2003subgradient} to solve (\ref{L*}) to approximate the optimal value $L^{*}$ gradually. 
	\begin{align}\label{L*}
		\max _{\boldsymbol{\mu}, \boldsymbol{\lambda}} &\sum_{l=0}^{L-1} \sum_{k=0}^{n_{l}} \tilde{f}_{l, k}\left(\mu_{k}^{l}, \lambda_{k}^{l}\right) +\sum_{l=1}^{L} f_{l}\left(\boldsymbol{\mu}^{l}, \boldsymbol{\lambda}^{l-1}\right)+f_{0}\left(\boldsymbol{\mu}^{0}\right) ,
	\end{align}	
	where $n_{l}$ is the size of layer $l$.
	We combine the original subgradient method with the Adam algorithm \cite{kingma2014adam} to achieve convergence effectiveness. The details of the algorithm are provided as Algorithm~\ref{algo1}. 
	$\alpha, \beta_1, \beta_2 \text{ and } \varepsilon$ are the hyperparameters of the Adam algorithm. $\alpha$ is the step size of the subgradient update. $\beta_1 \text{ and } \beta_2$ are the exponential decay rates. $\varepsilon$ is the parameter to avoid the divisor becoming zero. $\boldsymbol{m} \text{ and } \boldsymbol{v}$ are the moment vectors of the Adam algorithm. $\boldsymbol{m}$ is the $1^{\rm{st}}$ moment vector. $\boldsymbol{v}$ is the $2^{\rm{nd}}$ moment vector. $\theta$ and $K$ are the parameters we set to stop the iteration. $\theta$ is the minimum threshold of error. $K$ is the total steps of iteration. $(\boldsymbol{x}, y)$ is the test example to be verified. $f$ is the trained classification hyperplane equation.
	
	\begin{algorithm}
		\caption{SDVer}
		\label{algo1}
		Input {  $\boldsymbol{x}, f, \alpha, \beta_1, \beta_2, \varepsilon, \theta, K$  }\\
		Output {$L\left(\boldsymbol{\mu}^{(k)}, \boldsymbol{\lambda}^{(k)} \right)$	}
		\begin{algorithmic}[1]
			
			\State Initialize $k = 0$, $\boldsymbol{\mu}^{l(0)} = 0 $, $\boldsymbol{m}_{\boldsymbol{\mu}^{l(0)}} = 0$,  $\boldsymbol{v}_{\boldsymbol{\mu}^{l(0)}} = 0$, 
			$\boldsymbol{\lambda}^{l(0)}=0$, 		$\boldsymbol{m}_{\boldsymbol{\lambda}^{l(0)}} = 0 $, 		$\boldsymbol{v}_{\boldsymbol{\lambda}^{l(0)}} = 0 $
			
			\While{$k < K$}
			\State	Calculate $\boldsymbol{x}^{l(k)}$, $\boldsymbol{z}^{l(k)}$ by minimizing  $L\left(\boldsymbol{\mu}^{(k)}, \boldsymbol{\lambda}^{(k)} \right)$ in (\ref{xz})
			
			\If{$L\left(\boldsymbol{\mu}^{(k)}, \boldsymbol{\lambda}^{(k)} \right) > 0$ or $\lvert   L\left(\boldsymbol{\mu}^{(k)}, \boldsymbol{\lambda}^{(k)} \right) - \hat{y} \cdot f(\boldsymbol{x}^{0(k)})  \rvert < \theta$}
			\State 	return $L\left(\boldsymbol{\mu}^{(k)}, \boldsymbol{\lambda}^{(k)} \right)$
			
			\Else
			\State	Calculate the subgradient of $L\left(\boldsymbol{\mu}, \boldsymbol{\lambda}\right)$ at $\boldsymbol{\lambda}^{l(k)}$ :
			\begin{align*}		
				g_{\boldsymbol{\lambda}^{l(k)}}=\boldsymbol{x}^{l+1(k)}-\boldsymbol{h}^{l}\left(\boldsymbol{z}^{l(k)}\right)
			\end{align*}
			\State	Calculate the subgradient of $L\left(\boldsymbol{\mu}, \boldsymbol{\lambda}\right)$ at $\boldsymbol{\mu}^{l(k)}$ : 
			\begin{align*}	
				g_{\boldsymbol{\mu}^{l(k)}}=\boldsymbol{z}^{l(k)}-W^{l} \boldsymbol{x}^{l(k)}-\boldsymbol{b}^{l} 
			\end{align*}	
			\If{$g_{\boldsymbol{\lambda}^{l(k)}}=g_{\boldsymbol{\mu}^{l(k)}}=0$}
			\State 	return $L\left(\boldsymbol{\mu}^{(k)}, \boldsymbol{\lambda}^{(k)} \right)$
			
			\Else
			\State Update $\boldsymbol{\lambda}^{l(k)}$:
			\begin{align}	
				&\boldsymbol{m}_{\boldsymbol{\lambda}^{l(k+1)}} \leftarrow \beta_1 \cdot \boldsymbol{m}_{\boldsymbol{\lambda}^{l(k)}} + (1-\beta_1) \cdot  \boldsymbol{g}_{\boldsymbol{\lambda}^{l(k)}}\nonumber\\[-0.1cm]
				&\boldsymbol{v}_{\boldsymbol{\lambda}^{l(k+1)}} \leftarrow \beta_2 \cdot \boldsymbol{v}_{\boldsymbol{\lambda}^{l(k)}} + (1-\beta_2) \cdot  \boldsymbol{g}^{2}_{\boldsymbol{\lambda}^{l(k)}}\nonumber\\[-0.1cm]
				&\boldsymbol{\hat{m}}_{\boldsymbol{\lambda}^{l(k+1)}} \leftarrow\boldsymbol{m}_{\boldsymbol{\lambda}^{l(k+1)}} /(1-\beta^{k+1}_1) \nonumber\\[-0.1cm]
				&\boldsymbol{\hat{v}}_{\boldsymbol{\lambda}^{l(k+1)}} \leftarrow\boldsymbol{v}_{\boldsymbol{\lambda}^{l(k+1)}} /(1-\beta^{k+1}_2) \nonumber\\[-0.1cm]
				&\boldsymbol{\lambda}^{l(k+1)} \leftarrow \boldsymbol{\lambda}^{l(k)} - \alpha \cdot \boldsymbol{\hat{m}}_{\boldsymbol{\lambda}^{l(k+1)}} / (\sqrt{\boldsymbol{\hat{v}}_{\boldsymbol{\lambda}^{l(k+1)}}} + \varepsilon )\nonumber
			\end{align}	
			
			\State	Update $\boldsymbol{\mu}^{l(k)}$:
			\begin{align}
				&\boldsymbol{m}_{\boldsymbol{\mu}^{l(k+1)}} \leftarrow \beta_1 \cdot \boldsymbol{m}_{\boldsymbol{\mu}^{l(k)}} + (1-\beta_1) \cdot  \boldsymbol{g}_{\boldsymbol{\mu}^{l(k)}}\nonumber	\\[-0.1cm]	
				&\boldsymbol{v}_{\boldsymbol{\mu}^{l(k+1)}} \leftarrow \beta_2 \cdot \boldsymbol{v}_{\boldsymbol{\mu}^{l(k)}} + (1-\beta_2) \cdot  \boldsymbol{g}^{2}_{\boldsymbol{\mu}^{l(k)}}\nonumber\\[-0.1cm]
				&\boldsymbol{\hat{m}}_{\boldsymbol{\mu}^{l(k+1)}} \leftarrow\boldsymbol{m}_{\boldsymbol{\mu}^{l(k+1)}} /(1-\beta^{k+1}_1)\nonumber\\[-0.1cm]				
				&\boldsymbol{\hat{v}}_{\boldsymbol{\mu}^{l(k+1)}} \leftarrow\boldsymbol{v}_{\boldsymbol{\mu}^{l(k+1)}} /(1-\beta^{k+1}_2)\nonumber\\[-0.1cm]		
				&\boldsymbol{\mu}^{l(k+1)} \leftarrow \boldsymbol{\mu}^{l(k)} - \alpha \cdot \boldsymbol{\hat{m}}_{\boldsymbol{\mu}^{l(k+1)}} / (\sqrt{\boldsymbol{\hat{v}}_{\boldsymbol{\mu}^{l(k+1)}}} + \varepsilon )\nonumber
			\end{align}
			
			\EndIf
			\EndIf
			
			\State	$k \leftarrow k + 1$
			
			\EndWhile
			
			\State return $L\left(\boldsymbol{\mu}^{(k)}, \boldsymbol{\lambda}^{(k)} \right)$
		\end{algorithmic}
	\end{algorithm}

	\section{Experiments and analysis}
	\subsection{Datasets}
	Experimental evaluation of SDVer is conducted on the MNIST \cite{lecun1998gradient} and the Fashion-MNIST (F-MNIST) \cite{xiao2017fashion} datasets.		
	MNIST is a widespread and standard dataset in the field of adversarial robustness evaluation. It consists of grayscale images of handwritten digits 0 to 9 with a pixel size of 28$\times$28, including 60000 training samples and 10000 test samples. F-MNIST is more challenging than MNIST for benchmarking machine learning algorithms. It consists of 10 categories of clothing images, with the same image size and the number of training and test samples as MNIST.
	
	\subsection{Experimental settings}
	\begin{table}[h]
		\begin{center}
			\begin{minipage}{\textwidth}
				\caption{Test accuracy of the trained SVMs with different kernels on the first 100 images of each test dataset}\label{tab2}
				\begin{tabular*}{\textwidth}{@{\extracolsep{\fill}}lcccccc@{\extracolsep{\fill}}}
					\toprule%
					& \multicolumn{2}{@{}c@{}}{MNIST} & \multicolumn{2}{@{}c@{}}{F-MNIST} \\\cmidrule{2-3}\cmidrule{4-5}%
					binary classifiers &  0 and 1  &  4 and 9 & ankle-boot and bag &  shirt and coat  \\
					\midrule
					linear kernel  & 100\% & 100\%  & 100\%  & 86\%\\
					2-polynomial kernel  & 100\%  &100\%   & 100\%    & 93\%\\
					3-polynomial kernel  & 100\% &100\%   & 100\% & 92\%\\
					RBF kernel  & 100\% & 100\%   & 100\%   & 92\%\\
					\botrule
				\end{tabular*}
			\end{minipage}
		\end{center}
	\end{table}
	
	We compared SDVer with SAVer \cite{ranzato2019SAVer}, a state-of-the-art robustness verification method for SVMs based on the abstract interpretation. The SAVer evaluated the adversarial robustness of one-versus-one (OVO) multi-class SVMs on the MNIST and F-MNIST datasets. The OVO multi-class SVM integrates 45 binary-class SVM. We focus on the adversarial robustness evaluation of binary-class SVMs. Binary classifiers trained on dissimilar and similar classes are chosen as robustness evaluation objects considering two extreme scenarios. In MNIST, we choose binary classifiers of dissimilar handwritten digits 0 and 1 and binary classifiers of similar handwritten digits 4 and 9 as robustness evaluation objects. In F-MNIST, we choose binary classifiers of dissimilar ankle-boot and bag and binary classifiers of similar shirt and coat as robustness evaluation objects. The parameters of the trained SVMs are consistent with those of the SVMs verified in SAVer. The robustness of the trained SVMs is evaluated on the first 100 images of each test dataset. Table~\ref{tab2} shows the test accuracy of the trained SVMs on the test set. The proposed SDVer is encoded with Pytorch and runs on an NVIDIA Geforce RTX 3090 64GB.
	
	The subgradient method with the Adam optimizer is used to solve the optimization problem (\ref{L*}). The step size of updating parameters is linearly decreased between the initial learning rate {$\alpha_1$} and the final learning rate {$\alpha_2$}. The values of {$\alpha_1$} and {$\alpha_2$} are different for different kernels of SVMs. We set $\alpha_1 \in \{10^{-5}, 10^{-4}, 10^{-3}, 10^{-2} \}$ and $\alpha_2 \in \{10^{-10}, 10^{-9}, 10^{-8}, 10^{-7}, 10^{-6} \}$. The maximum value of $K$ is 10350000. We set other hyperparameters $\beta_1=0.9, \beta_2 = 0.999, \varepsilon = 10^{-8} \text{ and } \theta = 0.001$ for all tasks. 
	
	\subsection{Results and analyse}
\begin{figure}[htbp]
		\centering
		\includegraphics[width=1.2\textwidth]{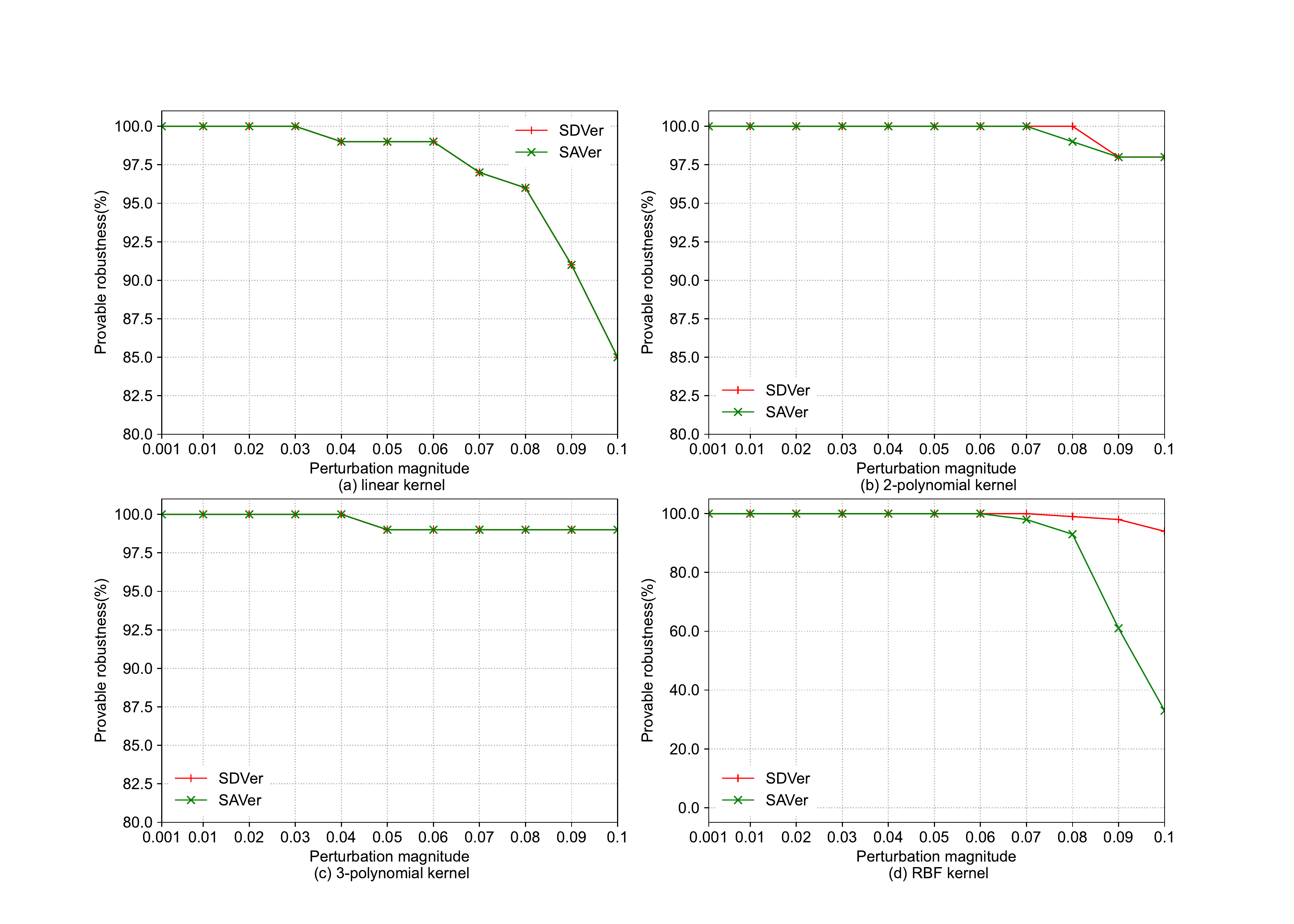}
		\caption{Provable robustness of handwritten digits 0 and 1 classifiers with different kernels}\label{fig1}
	\end{figure}

\begin{figure}[htbp]
		\centering
		\includegraphics[width=0.75\textwidth]{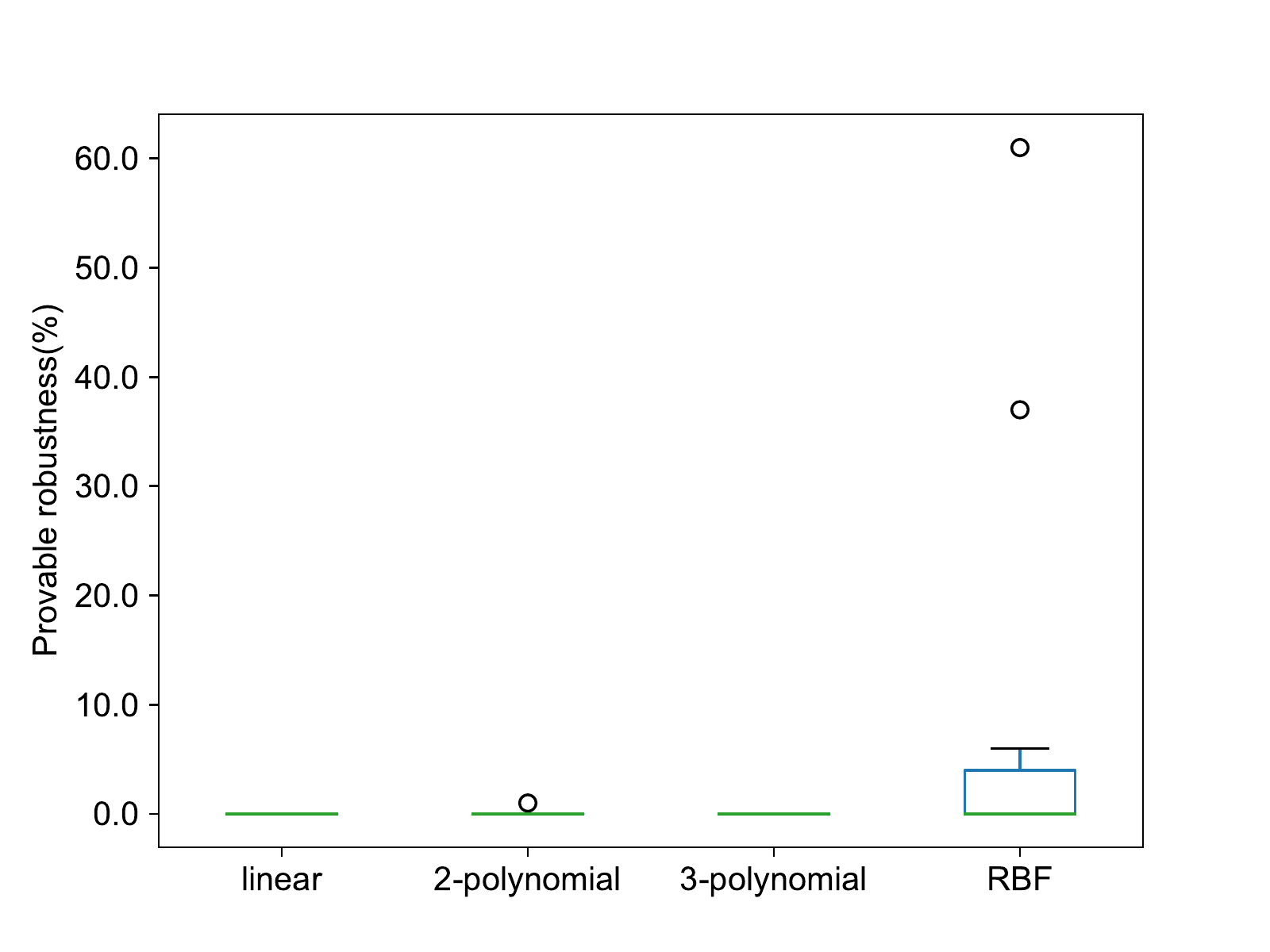}
		\caption{Provable robustness differences from SDVer to SAVer in Fig.~\ref{fig1}}\label{fig1-box}
	\end{figure}

\begin{figure}[htbp]
		\centering
		\includegraphics[width=1.2\textwidth]{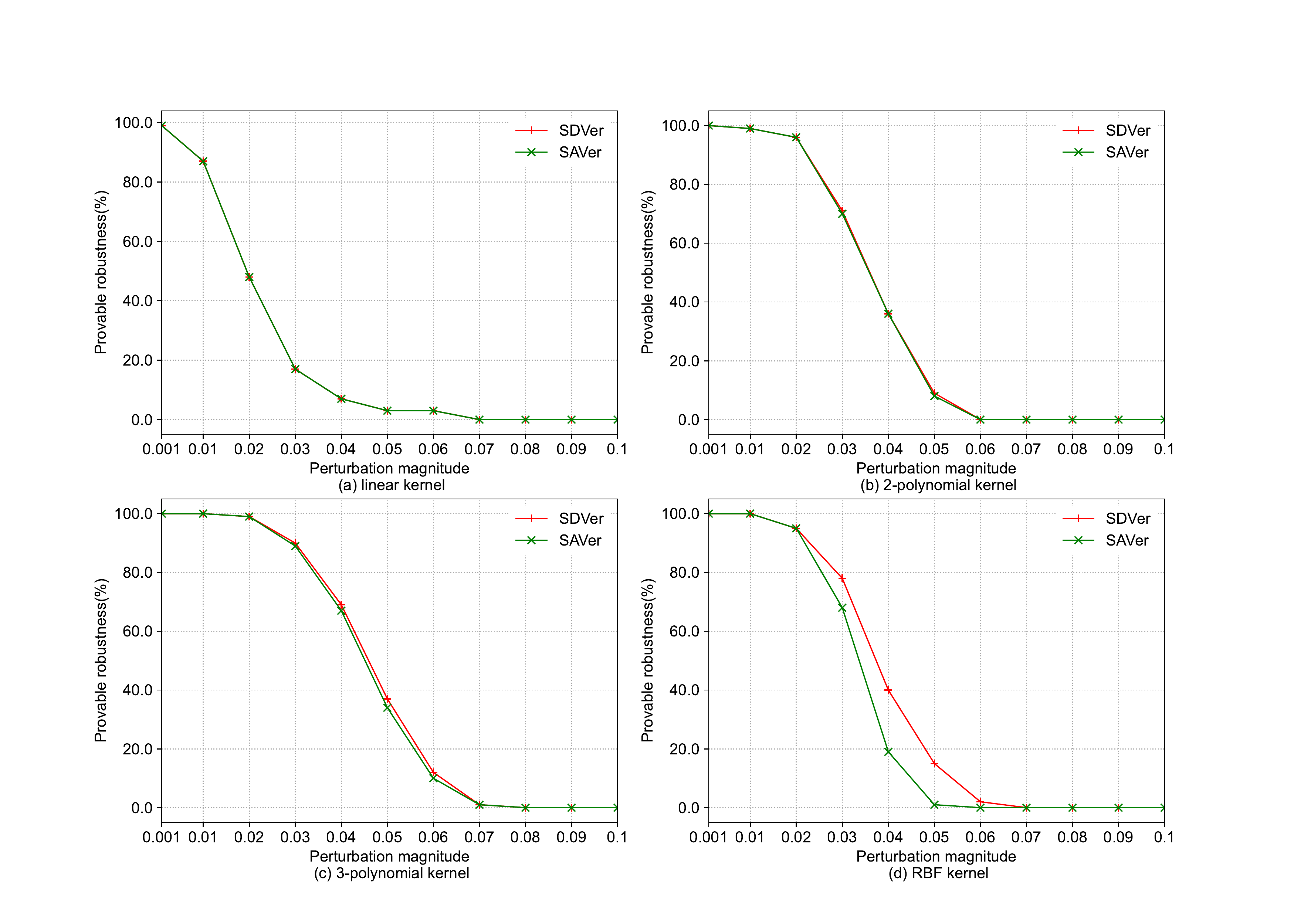}
		\caption{Provable robustness of handwritten digits 4 and 9 classifiers with different kernels}\label{fig2}
	\end{figure}

\begin{figure}[htbp]
		\centering
		\includegraphics[width=0.75\textwidth]{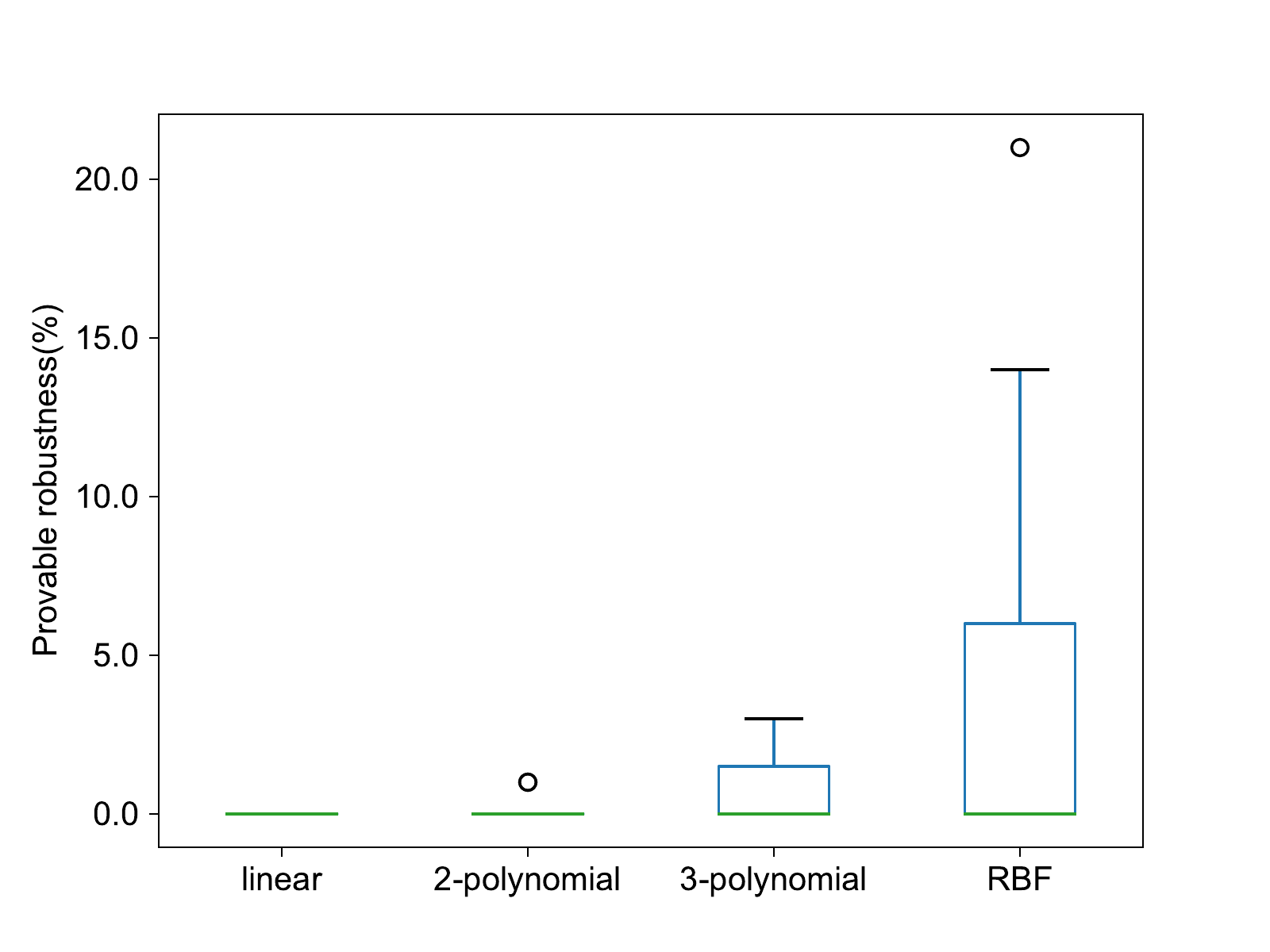}
		\caption{Provable robustness differences from SDVer to SAVer in Fig.~\ref{fig2}}\label{fig2-box}
	\end{figure}

\begin{figure}[htbp]
		\centering
		\includegraphics[width=1.2\textwidth]{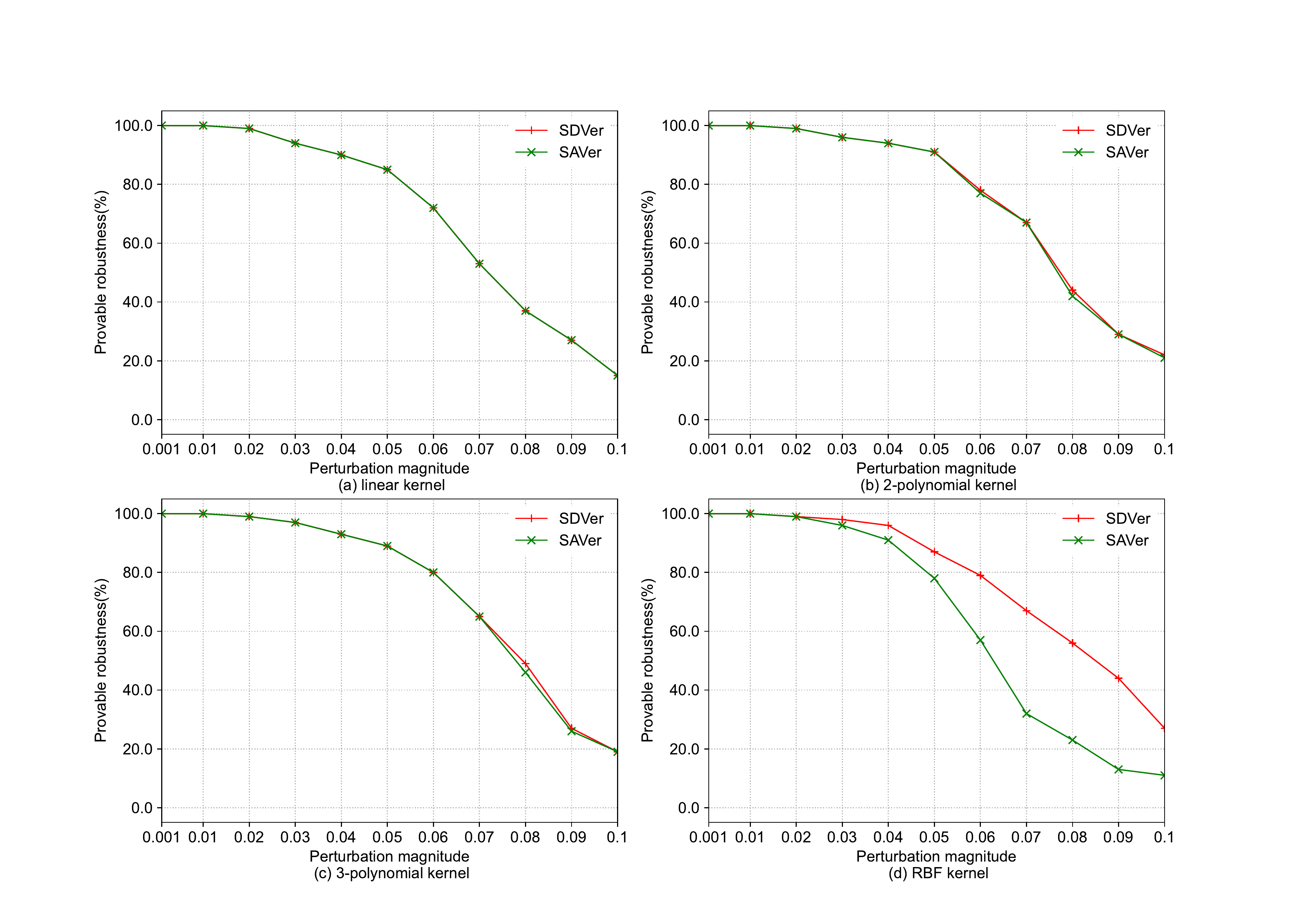}
		\caption{Provable robustness of ankle-boot and bag classifiers with different kernels}\label{fig3}
	\end{figure}

\begin{figure}[htbp]
		\centering
		\includegraphics[width=0.75\textwidth]{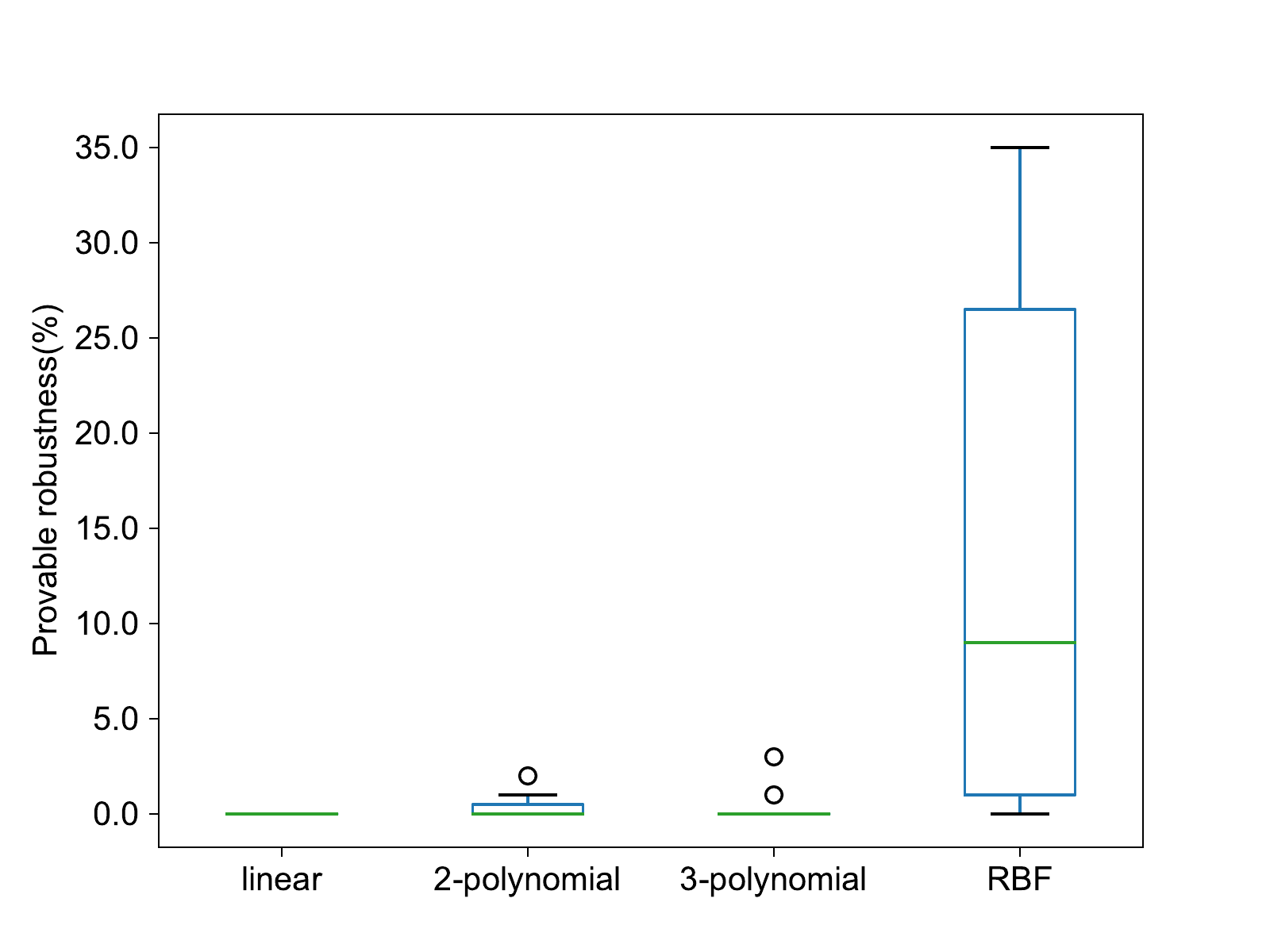}
		\caption{Provable robustness differences from SDVer to SAVer in Fig.~\ref{fig3}}\label{fig3-box}
	\end{figure}

\begin{figure}[htbp]
		\centering
		\includegraphics[width=1.2\textwidth]{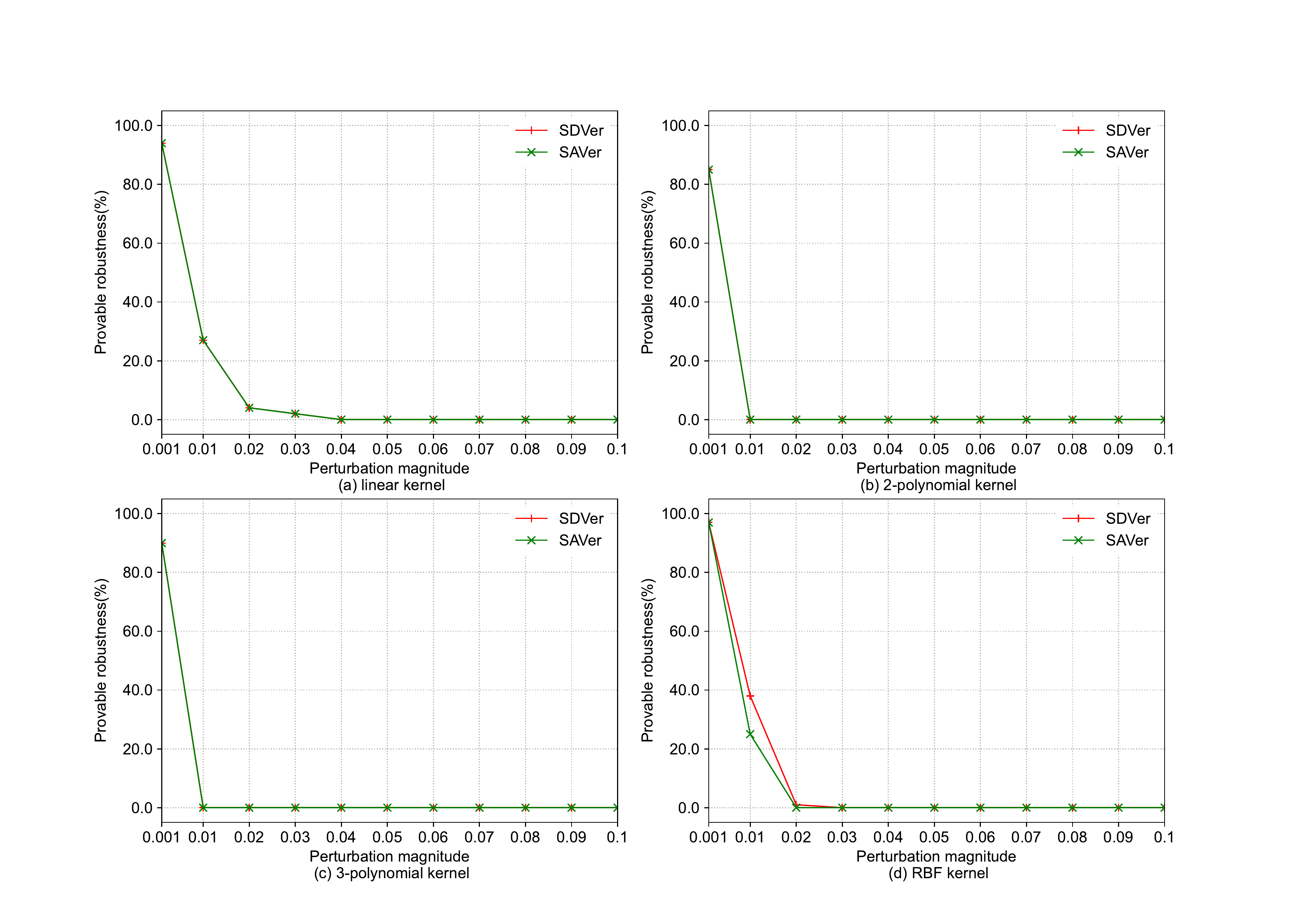}
		\caption{Provable robustness of shirt and coat classifiers with different kernels}\label{fig4}
	\end{figure}

\begin{figure}[htbp]
		\centering
		\includegraphics[width=0.75\textwidth]{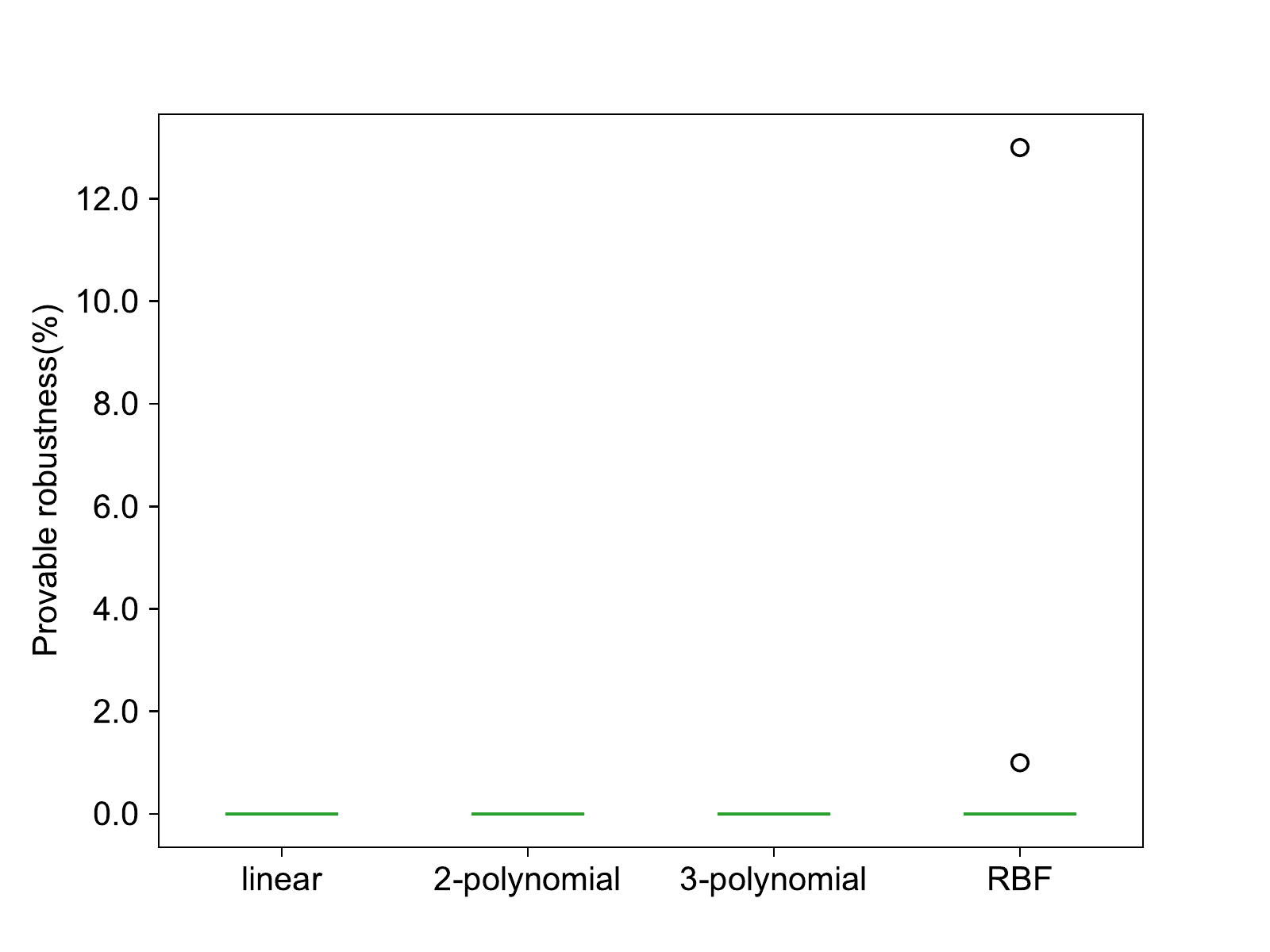}
		\caption{Provable robustness differences from SDVer to SAVer in Fig.~\ref{fig4}}\label{fig4-box}
	\end{figure}

	We are interested in the percentage of provable robustness, that is, the fraction of the test set that is robust. Figures~\ref{fig1} -~\ref{fig2-box} are the experimental results on the MNIST dataset. The robustness evaluation objects in Figs.~\ref{fig1} and~\ref{fig1-box} are  classifiers for handwritten digits 0 and 1. The robustness evaluation objects in Figs.~\ref{fig2} and~\ref{fig2-box} are classifiers for handwritten digits 4 and 9. Figures~\ref{fig3} -~\ref{fig4-box} are the experimental results on the F-MNIST dataset. The robustness evaluation objects in Figs.~\ref{fig3} and~\ref{fig3-box} are classifiers for ankle-boot and bag. The robustness evaluation objects in Figs.~\ref{fig4} and~\ref{fig4-box} are classifiers for shirt and coat.
	
	Figure~\ref{fig1} shows the provable robustness of handwritten digits 0 and 1 classifiers with different kernels evaluated with SDVer and SAVer. Figure~\ref{fig1-box} shows a box-plot of the provable robustness differences from SDVer to SAVer in Fig.~\ref{fig1}.		
	Figures~\ref{fig1}(a) and~\ref{fig1-box} show that our method obtains the same percentage of provable robustness as SAVer for SVMs with linear kernels.		
	Figures~\ref{fig1}(b) and~\ref{fig1-box} show that the evaluation result of our method is slightly better than that of SAVer for SVMs with 2-polynomial kernels. 		 
	Figures~\ref{fig1}(c) and~\ref{fig1-box} show that our method obtains the same percentage of provable robustness as SAVer for SVMs with 3-polynomial kernels.		
	Figures~\ref{fig1}(d) and~\ref{fig1-box} show that our method demonstrates a significantly better percentage of provable robustness than SAVer for SVMs with RBF kernels.	
	
	Figure~\ref{fig2} shows the provable robustness of handwritten digits 4 and 9 classifiers with different kernels evaluated with SDVer and SAVer.	
	Figure~\ref{fig2-box} shows a box-plot of the provable robustness differences from SDVer to SAVer in Fig.~\ref{fig2}.	
	Figures~\ref{fig2}(a) and~\ref{fig2-box} show that our method obtains the same percentage of provable robustness as SAVer for SVMs with linear kernels. 		
	Figures~\ref{fig2}(b),~\ref{fig2}(c), and~\ref{fig2-box} show that our method achieves a slightly better percentage of provable robustness than SAVer for SVMs with polynomial kernels. 		
	Figures~\ref{fig2}(d) and~\ref{fig2-box} show that our method demonstrates a significantly better percentage of provable robustness than SAVer for SVMs with RBF kernels. 		
	Note that the percentage of provable robustness for the handwritten digits 4 and 9 classifiers is lower than that of handwritten digits 0 and 1 classifiers in the same adversarial region. The reason is that handwritten digits 4 and 9 are very similar, 
	so handwritten digits 4 and 9 classifiers are more susceptible to adversarial perturbations.
	
	Figure~\ref{fig3} shows the provable robustness of ankle-boot and bag classifiers with different kernels evaluated with SDVer and SAVer.
	Figure~\ref{fig3-box} shows a box-plot of the provable robustness differences from SDVer to SAVer in Fig.~\ref{fig3}.		
	The conclusions drawn from Figs.~\ref{fig3} and~\ref{fig3-box} are consistent with those drawn from Figs.~\ref{fig2} and~\ref{fig2-box}.
	Note that there is a clear difference between ankle-boot and bag. The ankle-boot and bag classifiers achieve 100\% accuracy on the test set. In similar cases, the percentage of provable robustness of the ankle-boot and bag classifiers is lower than that of the handwritten digits 0 and 1 classifiers.
	The reason for this may be that ankle-boot and bag images carry more information than handwritten digits 0 and 1 images, making them more susceptible to adversarial perturbations.
	
	Figure~\ref{fig4} shows the provable robustness of shirt and coat classifiers with different kernels evaluated with SDVer and SAVer.
	Figure~\ref{fig4-box} shows a box-plot of the provable robustness differences from SDVer to SAVer in Fig.~\ref{fig4}.				
	Figures~\ref{fig4}(a),~\ref{fig4}(c), and~\ref{fig4-box} show that our method obtains the same percentage of provable robustness as SAVer for SVMs with linear and polynomial kernels. Figures~\ref{fig4}(d) and~\ref{fig4-box} show that our method demonstrates a higher percentage of provable robustness than SAVer for SVMs with RBF kernels. The accuracy of the shirt and coat classifiers on the test set is not very high. In addition, the difference between the shirt and coat is not noticeable. These reasons may lead to a lower percentage of provable robustness of the shirt and coat classifiers.
	
	\section{Conclusion}
	In this paper, SDVer is proposed for evaluating the adversarial robustness of SVMs. The method is based on the Lagrangian duality considering the kernels used in SVMs. The proposed SDVer is a robustness verification method that provides provable robustness for SVMs against various adversarial attacks. The state-of-the-art method for robustness verification of SVMs is SAVer, which uses an abstraction that combines interval domains and reduced affine form domains. When robustness evaluations are performed on SVMs with nonlinear kernels, the abstract nonlinear operations applied by SAVer on the interval domain and the reduced affine form domain lead to a loss of computational accuracy. Unlike SAVer, our method directly solves the verification problem using optimization techniques that improve the percentage of provable robustness of SVMs with nonlinear kernels. We conducted experiments on the MNIST and F-MNIST datasets. The proposed method obtained the same robustness evaluation results as SAVer in the case of linear kernels. The proposed method achieved better robustness evaluation results than SAVer for nonlinear kernels.	Our method can be expected to be applied in safety-critical fields such as malware detection, intrusion detection, and spam filtering. For now, the time complexity of our proposed method is sublinear, while the time complexity of the SAVer is linear. Our method has no advantage in terms of time consumption. How reduce the time complexity of our proposed method is a challenging direction.	Our proposed method is suitable for evaluating the adversarial robustness of binary-class SVMs. The multi-class SVMs are also widely used in safety-critical fields. In the future, we also will extend our robustness verification method for binary-class SVMs to multi-class SVMs.
	
	\section*{Declarations}
	
	\begin{itemize}

		\item Conflict of interest
		
		The authors declare that they have no conflict of interest.

		\item Availability of data and materials
		
		The data used in this paper are all from public datasets.

	\end{itemize}
	
	\bibliography{sn-bibliography}
	
\end{document}